\documentclass[conference]{IEEEtran}
\IEEEoverridecommandlockouts
\usepackage{url}
\usepackage{cite}
\usepackage{amsmath,amssymb,amsfonts}
\usepackage{algorithmic}
\usepackage{graphicx}
\usepackage{textcomp}
\usepackage{stfloats}
\usepackage{xcolor}
\usepackage{array}
\usepackage{booktabs}
\usepackage{makecell}
\usepackage{hyperref}
\usepackage{float}
\usepackage{multirow}
\usepackage{booktabs}

\def\BibTeX{{\rm B\kern-.05em{\sc i\kern-.025em b}\kern-.08em
    T\kern-.1667em\lower.7ex\hbox{E}\kern-.125emX}}

\newcommand{\ie}{\mbox{{\em i.e.}}}
\newcommand{\etal}{\textit{et al.}}

\begin{document}

\title{Seeing What the Vehicle Sees: Video-Augmented Virtual Reality for Physical Autonomous Vehicles}

\author{
\IEEEauthorblockN{
Md Tanjemul Islam,
Mohammad Shafin, and
Md Rafiul Kabir
}
\IEEEauthorblockA{
School of Engineering and Technology, Central Michigan University, Mount Pleasant, MI, USA\\
Email: \{tanje2m, shafi2m, kabir2m\}@cmich.edu
}
}
\maketitle

\begin{abstract}
Autonomous vehicles are expected to improve road safety and efficiency, but passengers often remain uncertain about what the vehicle perceives and why it acts as it does. Virtual reality (VR) offers a safe and repeatable medium for presenting this information, yet most passenger-facing VR studies rely on fully simulated vehicles or pre-scripted scenarios, so the motion and perception shown to the user do not originate from a physically operating autonomous system. This paper presents a video-augmented VR framework that couples a physical ROS 2 autonomous robot vehicle to a Unity 6 application deployed on a Meta Quest 3S headset. The vehicle state and live onboard camera stream are transmitted over two independent communication channels, allowing the virtual vehicle to mirror the physical robot's motion while the passenger simultaneously views the vehicle's first-person camera feed and its navigation decisions through an in-vehicle dashboard interface. We evaluate the framework over 20 repeated closed-loop navigation trials. The system achieves a mean state-update latency of 29.63 ms, a mean relative route-progress error of 2.28\% between the physical and virtual vehicles, and video delivery at 10.006 frames per second with 0.25\% frame loss. All monitored navigation decisions were correctly reflected in the VR interface with no missed or incorrect notifications. The results indicate that the framework can support temporally synchronized, semantically consistent, and accurate route-progress representation for immersive observation of physical autonomous-vehicle behavior.
\end{abstract}


\section{Introduction}
Autonomous vehicles are expected to improve transportation safety and efficiency, yet the passenger experience remains an important challenge. Unlike a human driver, an automated system does not naturally communicate what it has perceived or why it has selected a particular maneuver. Consequently, actions such as sudden braking or deceleration may feel abrupt or confusing when their underlying cause is not apparent to the passenger. Providing timely information about the vehicle’s perception and decisions can therefore make automated driving more transparent and easier to understand \cite{omeiza2021explanations}.
Virtual reality (VR) provides a safe, repeatable, and controlled medium for presenting such information \cite{hock2017carvr}. It can recreate driving scenarios and convey the vehicle’s surroundings, decisions, and movements without exposing participants to real traffic. VR also offers a practical means of observing autonomous-vehicle operation when direct access to a full-scale platform is limited, making it useful for demonstration and instructional purposes. However, most existing passenger-facing VR systems rely on fully simulated vehicles or pre-programmed scenarios \cite{serrano2021realistic}. In these systems, both the vehicle motion and the displayed perception are generated within the simulation rather than originating from a physically operating autonomous platform. Consequently, the behavior presented in VR may not reflect the sensing, decisions, and motion of a real autonomous vehicle.

This paper presents a video-augmented VR framework that connects a physical Robot Operating System (ROS) 2 autonomous robot vehicle to a virtual environment deployed on a Meta Quest 3S headset. The physical vehicle performs camera-based lane following and traffic-sign recognition, and transmits its position, heading, speed, acceleration, lane state, sign detections, and resulting navigation decisions over a dedicated User Datagram Protocol (UDP) channel. A second, independent UDP channel carries the live onboard camera stream. In the headset, the virtual vehicle mirrors the motion of the physical robot while an in-vehicle dashboard presents the live camera feed alongside the vehicle's current decision, so the passenger observes what the vehicle sees, what it decides, and how it moves in real-time. A video demonstration of the operation is available online~\cite{islam2026video_demo}.

The key contributions of this paper are:
\begin{itemize}
    \item A video-augmented VR framework coupling a physical ROS 2 autonomous vehicle to an immersive passenger view that combines synchronized motion, live onboard video, and navigation decisions.
    \item A dual-channel UDP architecture transmitting vehicle state and video independently to maintain real-time physical–virtual consistency.
    \item An experimental characterization over 20 closed-loop trials quantifying synchronization accuracy, latency, frame delivery, and decision-display fidelity.
\end{itemize}
\section{Related Work}
Prior work relevant to this paper falls into three areas: passenger-facing
visualization of autonomous-vehicle behavior, mixed-reality driving systems
incorporating physical vehicles~\cite{hock2017carvr}, and real-time
synchronization between physical robots and virtual
environments~\cite{yun2021virtualization}. These have largely been pursued
separately.
Passenger-facing interfaces have been introduced to share details about
an autonomous vehicle's environment, its current state, and its planned
actions~\cite{zou2021road, goedicke2022xr, mcgill2022passengxr}. Morra
\etal~developed an immersive VR driving simulator containing a head-up display
that presented visual cues from the vehicle's sensory and planning
systems~\cite{morra2019building}. The interface presented detected objects,
traffic signs, traffic lights, navigation information, warning conditions, and
additional elements with the potential to influence vehicle behavior. The
results underscored the significance of providing passengers with pertinent
information regarding the vehicle and its environment. However, the vehicle,
perception system, and driving scenarios were fully simulated. As a result, the
displayed information did not originate from a physically deployed autonomous
vehicle operating in real-world conditions. 

\begin{figure*}
    \centering
    \includegraphics[width=1.8\columnwidth]{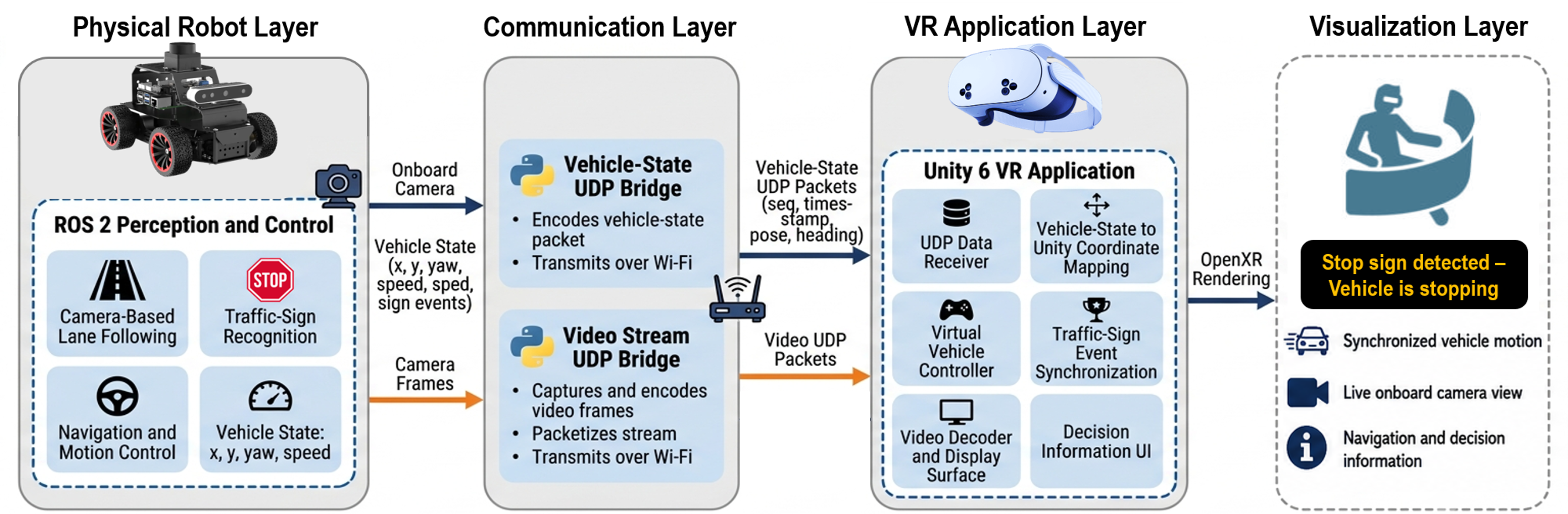}
\caption{Architecture of the proposed video-augmented VR system.}
    \label{fig:arch}
\end{figure*}

Beyond passenger information displays, VR has also been employed as an
exploratory medium for non-experts to examine autonomous-vehicle behavior; Owoputi \etal~\cite{owoputi2023ive2} developed IVE, an immersive virtual environment that enables hands-on exploration of security vulnerabilities in automotive ranging sensors without access to a physical
vehicle. This work likewise relied on a simulated vehicle rather than a physically operating platform.
Goedicke \etal~introduced XR-OOM, a Unity-based mixed-reality driving system
that combines a real vehicle with a video-passthrough headset~\cite{goedicke2022xr}. During actual driving, the platform aligns the
physical car, the participant's perspective, and virtual elements. Bu
\etal~later extended this approach through Portobello, which uses ROS-based
LiDAR localization to update a corresponding virtual vehicle in
Unity~\cite{bu2024portobello}. Related vehicle-in-the-loop studies have also synchronized physical
vehicles with virtual environments for immersive visualization and
system evaluation~\cite{weiss2022high}.

Liu \etal~presented a mixed-reality framework that synchronizes physical
ROS-based mobile robots with their virtual representations in
Unity~\cite{liu2020mobile}. Data regarding sensors,
positions, and orientations from the actual robots were employed to refresh the
virtual counterparts instantaneously. Although this study is technically
similar to the suggested ROS--Unity communication framework, its emphasis was
on simulating multiple robots instead of visualizing autonomous-vehicle
behavior for passengers. Overall, previous studies have examined passenger information interfaces, video-augmented mixed reality, and ROS--Unity synchronization separately. Few integrate live onboard video, vehicle motion, traffic-sign detections, and action information from a physical autonomous vehicle into one immersive VR application. This work addresses that gap.

\section{Framework Architecture and Implementation}
\subsection{Architecture Overview} Our proposed framework has four functional layers. Fig.~\ref{fig:arch} shows the architecture.
\subsubsection*{Physical Robot Layer}
The physical robot layer consists of a ROS 2–based autonomous robot car. Its perception tasks are primarily navigation and traffic sign recognition. It processes odometry data (\ie, position, orientation, speed), lane position, traffic sign information, and robot action-reason-related information, and sends it to the next layer. Also, the robot captures the first-person camera view, which is transmitted to the VR application.

\subsubsection*{Communication Layer}
This layer is responsible for proper communication of the data to its recipient. For convenience, robot state information and live video data are transmitted through two independent UDP communication channels. Separating telemetry from video allows data flow to be handled independently.
\subsubsection*{VR Application Layer}
This layer consists of the Unity-based VR application, which is deployed on the Meta Quest 3S headset. It receives robot state and video streams from the communication layer and processes them. The application synchronizes the virtual robot motion with the physical robot state, renders the virtual environment, and generates the navigation information interface.
\subsubsection*{Immersive Visualization Layer}
This is our final layer. It provides the user/passenger with a real-time immersive view of the autonomous robot's operation by integrating the virtual environment. It is synchronized with the physical robot car's motion, live onboard video, and decision information. There is a dashboard screen inside the virtual car that shows all information in a unified way.
\subsection{Implementation}
\subsubsection*{Robot State Processing and Transmission}
The robot state data were obtained from the ROS 2 navigation system. We implemented a UDP communication module to package and transmit timestamp, position, orientation, speed, and decision-related information to the VR headset. The continuous transmission enables real-time synchronization between the physical robot and the virtual representation.

\subsubsection*{Video Streaming Implementation}
We captured the robot's onboard camera stream and processed it separately. A dedicated UDP channel was used for video streaming. The robot's onboard camera captured RGB frames of $640\times480$ pixels. 
Each frame was JPEG encoded with a quality factor of 65 and transmitted as a single UDP datagram over a 2.4\, GHz Wi-Fi network. Unity received it and decoded the frames for display in the VR dashboard at a target rate of 10 FPS.

\begin{figure}
    \centering
    \includegraphics[width=\columnwidth]{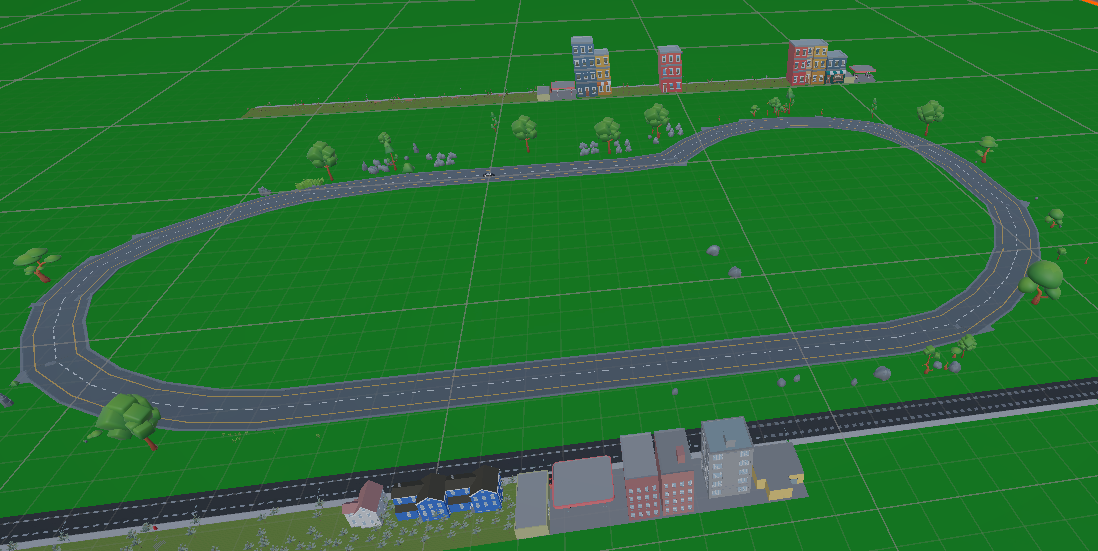}
    \caption{Unity-based virtual driving environment, shadowing the real physical autonomous vehicle system.}
    \label{fig:unity_map}
\end{figure}

\subsubsection*{Virtual Environment Design}
The virtual environment was developed in Unity 6.2 and is shown in Fig.~\ref{fig:unity_map}. The scene reproduces the closed-loop road structure of the physical test environment, together with a virtual model of the robot vehicle and the traffic signs along the route. Additional landscape elements were placed outside the driving surface to improve immersion; these are purely visual and do not affect synchronization. Because the ROS and Unity coordinate conventions differ, the two environments were spatially aligned before the trials: a mapping ratio was established from the measured road dimensions, the robot's planar coordinates were mapped onto the Unity X--Z plane with the corresponding axis and heading corrections, and the alignment was verified at several reference points along the route.

\subsubsection*{VR Application Deployment}
\begin{figure}
    \centering
    \includegraphics[width=.89\columnwidth]{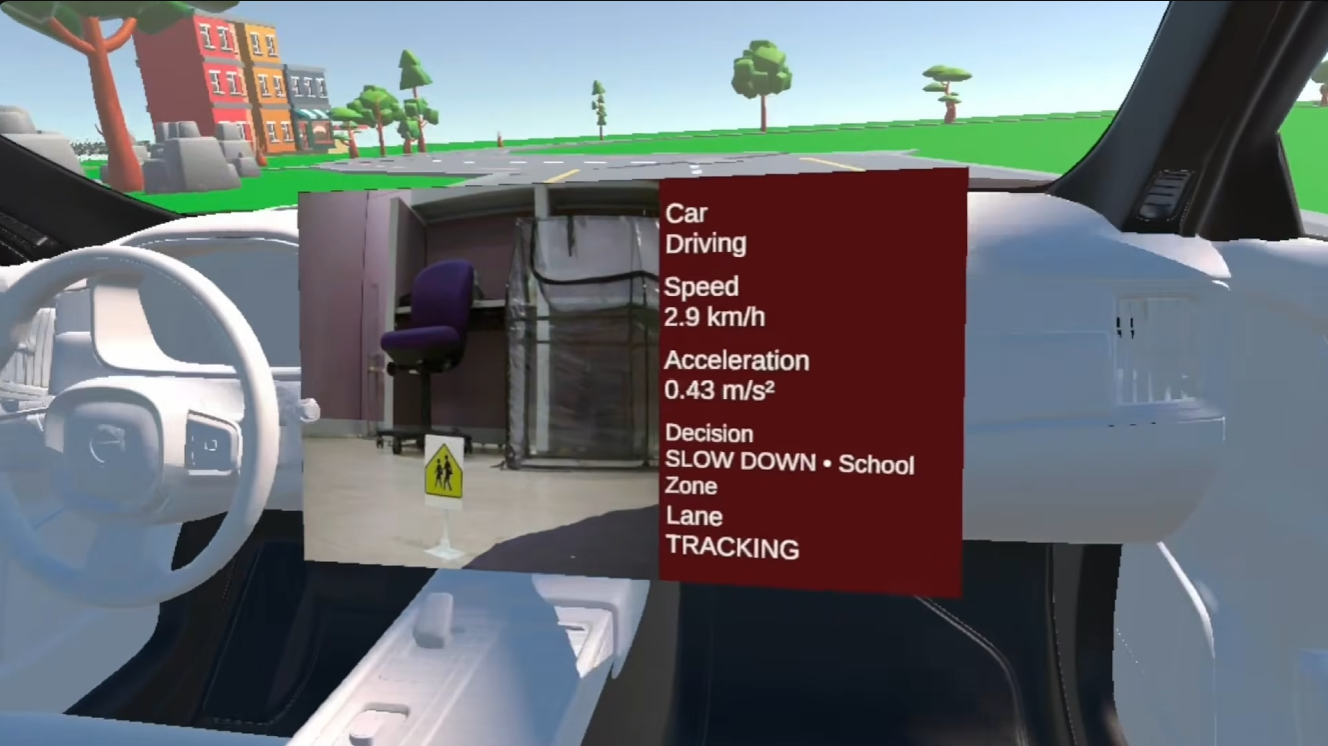}
    \caption {First-person view of the passenger in the VR realm. }
    \label{fig:vr_cockpit}
\end{figure}

The Unity application was deployed on a Meta Quest 3S headset as a standalone
Android build. The headset receives the robot state and video streams directly
and renders the synchronized virtual robot, the live camera feed, and the
navigation information, while six-degree-of-freedom tracking allows the
passenger to look around the virtual cabin naturally. Fig.~\ref{fig:vr_cockpit}
shows the passenger's front-seat view: the physical vehicle's live video stream
is presented on an in-cabin dashboard screen, with real-time behavioral
information along its right edge.

\section{Experimental Setup and Results}

\subsection{Setup and Evaluation Metrics}
We experimented with the implemented system to evaluate the effectiveness of the proposed framework, analyzing synchronization between the physical robot and the VR representation, live video transmission, and visualization of autonomous navigation decisions. During the experiments, the Unity application ran on a workstation, which received the video stream. The Raspberry Pi and the VR workstation desktop were synchronized to the campus NTP server before each run; the measured offset was found to be approximately 1\,ms, which is negligible. The physical robot followed a predefined closed-loop path and generated navigation information including robot pose, velocity, acceleration, lane position, traffic-sign detection, and the corresponding navigation decisions. The VR application synchronized the virtual robot motion with the physical robot, displayed the live camera feed, and presented navigation-related information through an in-vehicle dashboard interface. The framework was evaluated under autonomous navigation scenarios containing stop, speed-limit, speed-advisory, railroad-crossing, and school-crossing signs. We conducted 20 repeated closed-loop trials under consistent conditions, each lasting approximately 70\,s, and recorded the robot state, VR state, transmitted packets, video frames, and displayed notifications throughout each run.
System performance was analyzed in terms of robot–VR synchronization, communication latency, video-streaming performance, and navigation visualization. The communication latency was calculated as the time difference between the transmission of robot state data and its replication in VR, as shown in \eqref{eq:latency}.

\begin{equation}
T_L = T_{exe} - T_{send}
\label{eq:latency}
\end{equation}

where $T_{send}$ represents the timestamp when the robot state data is
transmitted, and $T_{exe}$ represents the timestamp when the VR application has executed the action.
The synchronization accuracy between the physical robot and the virtual robot was evaluated using the progress error equation \eqref{eq:progress_error}.

\begin{equation}
E_p = \frac{|P_r - P_v|}{P_r} \times 100
\label{eq:progress_error}
\end{equation}

where $P_r$ represents the physical robot progress along the predefined path and $P_v$ represents the virtual robot progress in the VR environment. 
For video-streaming latency, when a packet was transmitted from the Raspberry Pi, a timestamp was recorded immediately before it was sent through the UDP channel. Upon receiving the packet at the workstation, a second timestamp was recorded. Thus, the reported latency represents the one-way UDP packet delivery latency between the Raspberry Pi and the workstation and does not include camera capture, video encoding, decoding, or rendering time.

\subsection{Robot--VR Synchronization Performance}
Communication latency was measured as the time required for robot state information to reach the VR application and update the virtual representation. As shown in Table~\ref{tab:synchronization_results}, the system achieved an average latency of 29.63\,ms with a standard deviation of 12.86\,ms and RMSE of 32.29\,ms. The average value corresponds to a state refresh well within the range required for continuous motion updates. This indicates that the dual-channel design delivers robot state to the VR application quickly enough to support real-time visualization. Synchronization accuracy was assessed by comparing the robot state transmitted from the physical platform against the corresponding state of the virtual robot in the VR environment. The system achieved an average relative progress error of 2.28\% with a standard deviation of 5.46\% and an RMSE of 0.48\%. The two representations (\ie, virtual and physical) remain closely aligned, indicating consistent tracking throughout autonomous navigation.

\begin{table}
\centering
\caption{Robot--VR Synchronization Performance}
\label{tab:synchronization_results}
\footnotesize
\setlength{\tabcolsep}{5pt}
\begin{tabular}{llc}
\toprule
\textbf{Category} & \textbf{Metric} & \textbf{Value} \\
\midrule
\multirow{3}{*}{Communication}
& Average Latency & 29.63 ms \\
& Latency Std. Deviation & 12.86 ms \\
& Latency RMSE & 32.29 ms \\
\midrule
\multirow{3}{*}{Synchronization}
& Average Relative Progress Error & 2.28\% \\
& Std. Deviation & 5.46\% \\
& Progress RMSE & 0.48\% \\
\bottomrule
\end{tabular}
\end{table}

\begin{figure}
    \centering
    \includegraphics[width=.96\columnwidth]{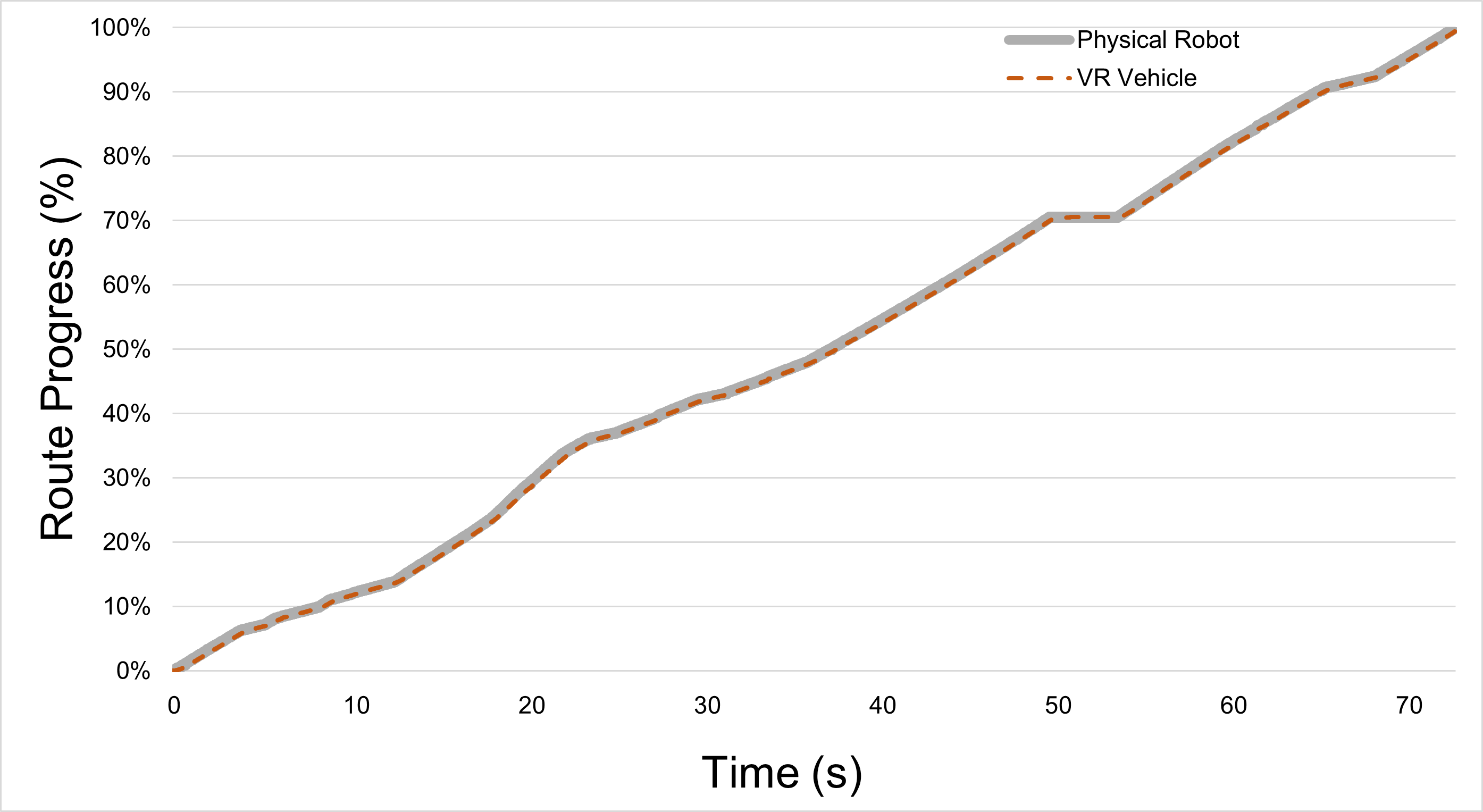}
    \caption{Physical robot and VR vehicle route progress over time.}
    \label{fig:Graph}
\end{figure}

Fig.~\ref{fig:Graph} compares the route progress of the physical robot and the corresponding virtual vehicle over time. The two trajectories closely overlap throughout the experiment, demonstrating consistent synchronization between the physical and virtual systems. Though minor deviations are observed during changes in robot speed and direction, mostly the lines follow each other closely.

\subsection{Video Streaming Evaluation}

The video streaming performance was evaluated to verify the capability of the proposed framework to deliver real-time camera visualization from the autonomous robot to the VR environment. The evaluation considered frame delivery reliability, streaming rate, and transmission latency.
Table~\ref{tab:video_streaming_results} summarizes the video-streaming performance. The system achieved a calculated frame rate of 10.006 FPS (frames per second), which closely matches the target frame rate of 10 FPS, with only 0.25\% frame loss during transmission. The average video latency was 6.03 ms, and the maximum observed latency was 152.89 ms, whereas recorded standard deviation was 3.85 ms. These results indicate that the UDP-based video communication channel can provide stable real-time camera streaming for immersive VR visualization.

\begin{table}
\centering
\caption{Video Streaming Performance}
\label{tab:video_streaming_results}
\footnotesize
\setlength{\tabcolsep}{5pt}
\begin{tabular}{llc}
\toprule
\textbf{Category} & \textbf{Metric} & \textbf{Value} \\
\midrule
\multirow{3}{*}{Frame Delivery}
& Target Frame Rate     & 10 FPS \\
& Measured Frame Rate   & 10.006 FPS \\
& Frame Loss            & 0.25\% \\
\midrule
\multirow{3}{*}{Latency}
& Average               & 6.03 ms \\
& Maximum               & 152.89 ms \\
& Std. Deviation        & 3.85 ms \\
\bottomrule
\end{tabular}
\end{table}

\subsection{Navigation Information Fidelity}
\begin{figure*}
    \centering
    \includegraphics[width=1.66\columnwidth]{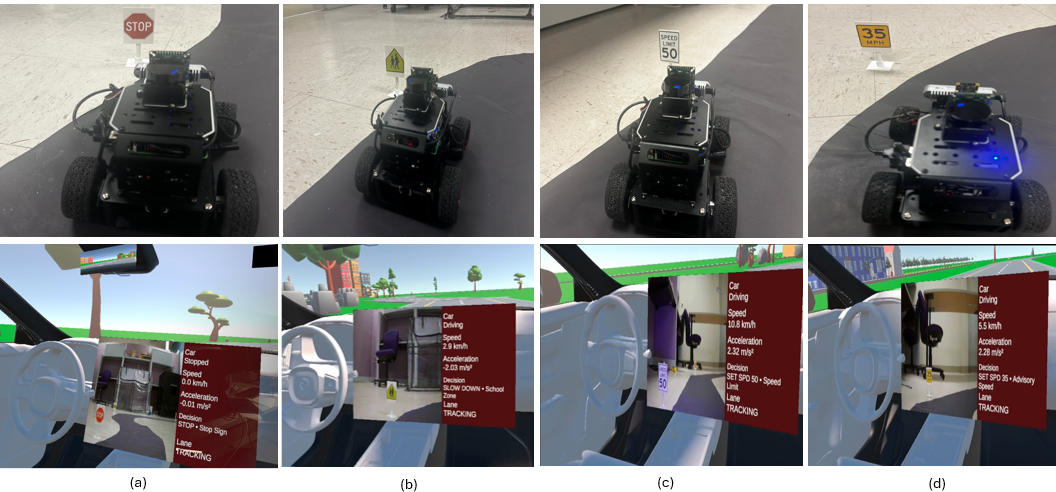}
    \caption{Representative physical autonomous-vehicle states and corresponding
    VR visualizations at different events: (a) stop-sign detection, (b) school-zone
    sign detection, (c) speed-limit sign detection, and (d)
    speed advisory sign detection.}
    \label{fig:vr3}
\end{figure*}

This module was validated by comparing the physical robot's notification information generation with VR dashboard information accuracy. Fig.~\ref{fig:vr3} presents five representative traffic-sign scenarios and their corresponding observations in the physical and VR environments. We identified the factors that alter the physical robot's behavior and assessed how accurately each was transmitted to the VR environment as action and display information. During the experiments, the robot encountered 5 different traffic signs that triggered predefined navigation actions. For each navigation action, the total number of occurrences across the 20 trials and the number of correctly synchronized display information were recorded. An event was considered successfully synchronized when the corresponding navigation information was received and displayed in the VR application without error or omission. Table~\ref{tab:decision_validation} summarizes the validation results. All events were correctly translated into the VR environment. Each physical observation is paired with its corresponding virtual observation to examine the consistency of vehicle behavior, traffic-sign information, and motion state between the physical and virtual platforms.

\begin{table}
\centering
\caption{Navigation Decision Visualization Validation}
\label{tab:decision_validation}
\footnotesize
\setlength{\tabcolsep}{3pt}

\begin{tabular}{lccc}
\toprule
\textbf{Item} &
\textbf{Robot Action} &
\textbf{VR Display} &
\textbf{VR Action Sync} \\
\midrule
Lane Position       & Lane Following & Lane Status   & \checkmark \\
Speed               & Speed Control  & Current Speed & \checkmark \\
Acceleration        & Motion Update  & Acceleration  & \checkmark \\
Stop Sign           & Vehicle Stop   & STOP          & \checkmark \\
Speed Limit 50      & Maintain Speed & SET SPD 50    & \checkmark \\
Speed Advisory 35   & Maintain Speed & SET SPD 35    & \checkmark \\
Railroad Crossing   & Reduce Speed   & SLOW DOWN     & \checkmark \\
School Crossing     & Reduce Speed   & SLOW DOWN     & \checkmark \\
\bottomrule
\end{tabular}
\end{table}

\section{Conclusion}
This paper presented a video-augmented VR framework that couples a physical autonomous robot vehicle to an immersive Unity 6 application deployed on a Meta Quest 3S headset. Vehicle state and the live onboard camera stream are carried over two independent UDP channels, allowing the virtual vehicle to mirror the motion of the physical robot while the passenger simultaneously observes the vehicle's first-person camera view and its navigation decisions through an in-vehicle dashboard. Across 20 closed-loop navigation trials, the virtual vehicle tracked the physical robot with a mean route-progress error of 2.28\%, with state updates and video frames delivered at latencies suitable for real-time observation, and all monitored navigation decisions were correctly reflected in the VR interface. These results indicate that the framework maintains progress, temporal, and semantic consistency required for real-time immersive observation of physical vehicle behavior.

Several directions remain. The present evaluation is technical; a human-subject study is needed to determine whether the combined video and decision display improves passenger awareness, predictability, and trust. Increasing the video frame rate and using hardware-accelerated encoding would bring the visual channel closer to passenger-facing responsiveness. Finally, the framework is unidirectional, and allowing passenger input to influence vehicle behavior would enable interactive studies of human–vehicle collaboration.

\bibliographystyle{IEEEtran}
\bibliography{mybib}
\end{document}